\documentclass[journal]{IEEEtran}
\usepackage{cite}
\usepackage{amsmath,amssymb,amsfonts}
\usepackage{algorithmic}
\usepackage{graphicx}
\usepackage{textcomp}
\def\BibTeX{{\rm B\kern-.05em{\sc i\kern-.025em b}\kern-.08em
    T\kern-.1667em\lower.7ex\hbox{E}\kern-.125emX}}
\usepackage{verbatim}
\usepackage{tabularx} 
\usepackage{booktabs} 
\usepackage{url}
\usepackage{enumitem}
\usepackage{algorithm2e}

\begin{document}
\title{Comparative Analysis and Ensemble Enhancement of Leading CNN Architectures for Breast Cancer Classification}
\author{Gary Murphy, Raghubir Singh, \IEEEmembership{Member, IEEE}
\thanks{
This research did not receive any specific grant from funding agencies in the public, commercial, or not-for-profit sectors.}
\thanks{G. Murphy, and R. Singh are with the University of Bath, Claverton Down, Bath, BA2 7AY, United Kingdom (e-mail:gpjm20@bath.ac.uk; rs3022@bath.ac.uk). }}

\maketitle

\begin{abstract}

This study introduces a novel and accurate approach to breast cancer classification using histopathology images. It systematically compares leading Convolutional Neural Network (CNN) models across varying image datasets, identifies their optimal hyperparameters, and ranks them based on classification efficacy. To maximize classification accuracy for each model we explore, the effects of data augmentation, alternative fully-connected layers, model training hyperparameter settings, and, the advantages of retraining models versus using pre-trained weights. Our methodology includes several original concepts, including serializing generated datasets to ensure consistent data conditions across training runs and significantly reducing training duration. Combined with automated curation of results, this enabled the exploration of over 2,000 training permutations - such a comprehensive comparison is as yet unprecedented. Our findings establish the settings required to achieve exceptional classification accuracy for standalone CNN models and rank them by model efficacy. Based on these results, we propose ensemble architectures that stack three high-performing standalone CNN models together with diverse classifiers, resulting in improved classification accuracy. The ability to systematically run so many model permutations to get the best outcomes gives rise to very high quality results, including  99.75\% for BreakHis x40 and BreakHis x200 and 95.18\% for the Bach datasets when split into train, validation and test datasets. The Bach Online blind challenge, yielded 89\% using this approach. Whilst this study is based on breast cancer histopathology image datasets, the methodology is equally applicable to other medical image datasets.

\end{abstract}

\begin{IEEEkeywords}
Breast cancer, Convolutional Neural Networks Comparison, Dataset serialization, Ensemble architectures, Reduced Training Duration
\end{IEEEkeywords}

\section{Introduction}
\label{sec:introduction}

\subsection{Breast Cancer}
Breast cancer globally is the most common form of cancer in women and the second most common form of any cancer \cite{b46}. It is estimated that there are 2.3m cases diagnosed globally each year \cite{b10}. It is also the leading cause of female death by cancer – 684,996 in 2020. \cite{b10}. It remains a significant global health challenge, with its impact and mortality partly mitigated by advances in screening and  treatment, particularly in developed countries.

Widespread adoption of screening programmes is seen as a critical step in lowering death rates in Breast Cancer in less developed countries and as such is a key goal of the Breast Health Global Initiative \cite{b88}. 
AI assisted Breast Cancer screening, offers potential in both developed and under-developed countries not just by improving accuracy, but in lowering the cost of detection through its efficiency.

This study focuses on the use of CNNs for breast cancer detection and classification based on histopathology image datasets. Through direct model comparison on different image datasets it seeks to determine which architecture(s) and which settings are most suitable for this task. 

\subsection{CNNs}
CNNs have revolutionised computer vision and classification, taking inspiration from the neural
architecture of the visual cortex in mammals’ vision \cite{b13}. Fukushima’s
groundbreaking neocognitron \cite{b12} together with early innovators such as LeCun
\cite{b15} set the foundations for the evolution of CNNs into the powerful models
available today.

\begin{figure}[!ht]
    \centering
    \includegraphics[width=0.8\linewidth]{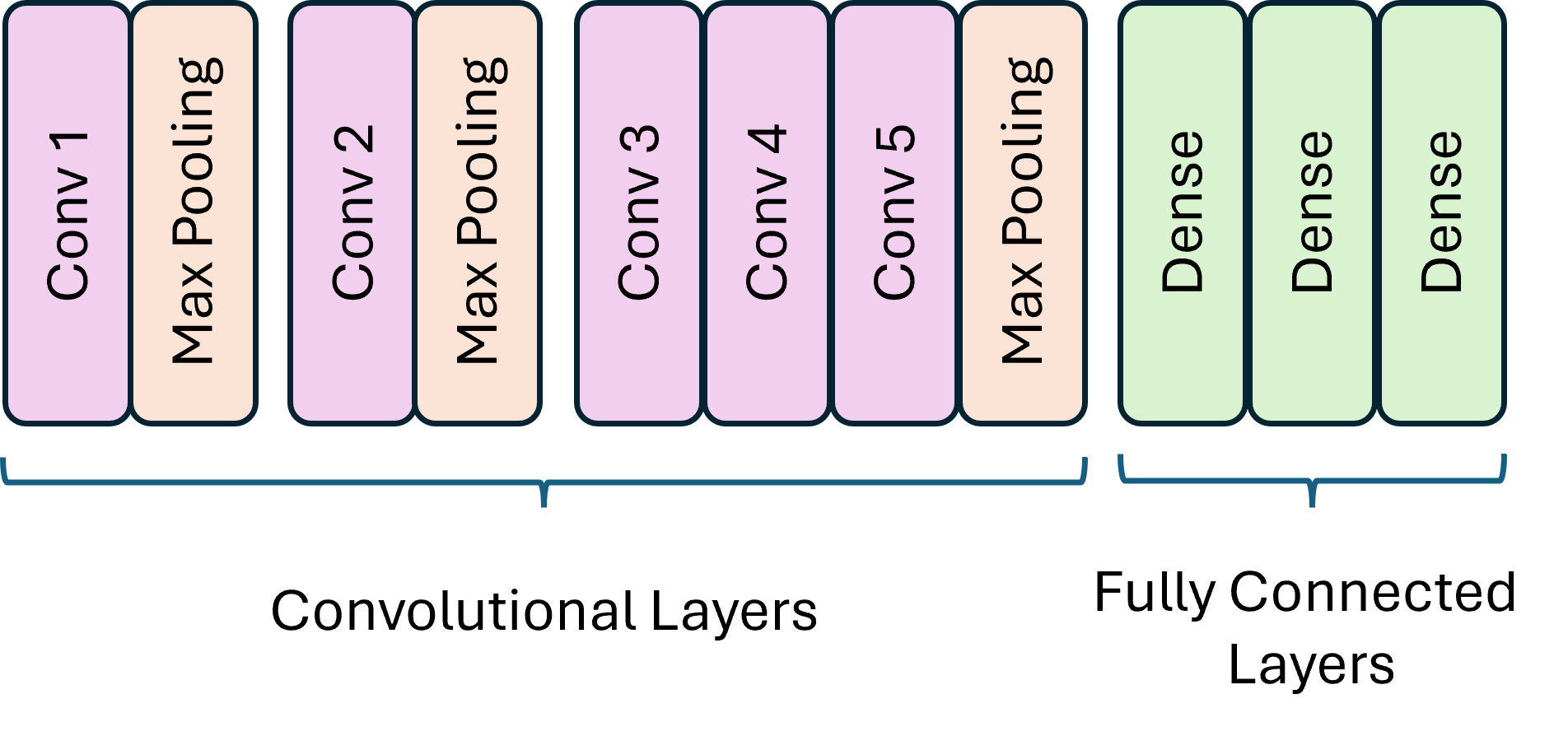}
    \caption{ Basic block architecture of AlexNet \cite{b22}}
    \label{fig:Example CNN}
\end{figure}

Since the inception of the ILSVRC \cite{b19}, and the pioneering breakthrough
in accuracy of AlexNet \cite{b22}, illustrated  in  Fig. \ref{fig:Example CNN}, also notable for exploiting GPUs for processing, increasing computer resources have permitted creation of deeper networks, such as the VGG models, VGG16 and VGG19 \cite{b23}. 

Subsequently CNNs evolved towards architectures prioritising efficiency and dealing with the problems surrounding ever deeper networks, such as vanishing gradients.

A raft of innovative CNN architectures have resulted, such as Inception \cite{b25},  ResNets \cite{b27} and other architectures that aimed to be more efficient, rather than just adding more and more layers. These include DenseNets \cite{b29}, Xception \cite{b53}, NASNets \cite{b32}, MobileNetV2 \cite{b31} and EfficientNets \cite{b28}. Variants of all these CNN models are fully optimised and evenly tested and compared against each other for breast cancer classification of histopathology images in this study.

\begin{figure}[!ht]
    \centering
    \includegraphics[width=1.0\linewidth]{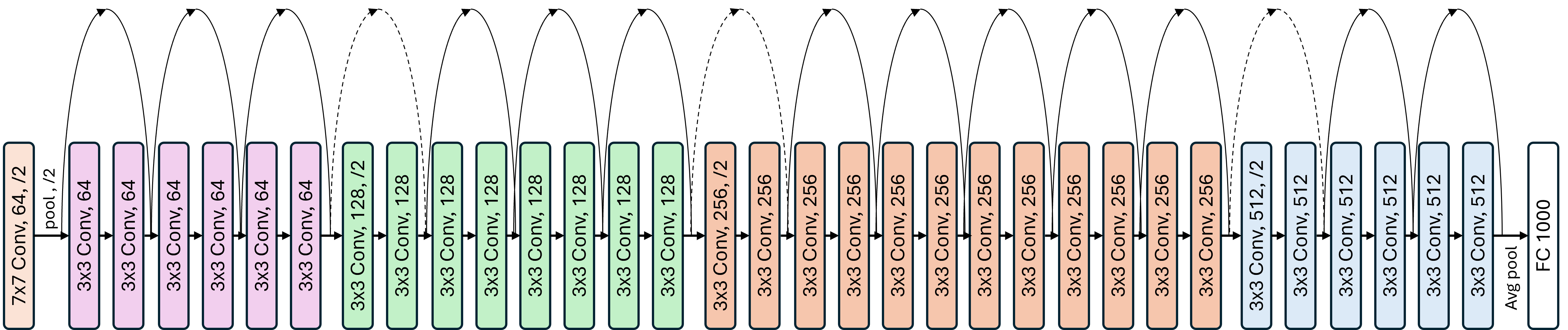}
    \caption{ResNet50 Simplified Block Architecture showing the 34 layers with learnable parameters.  The arced lines show the 'shortcuts' introduced in the model to enable its residual capability. Redrawn and transposed from vertical to horizontal \cite{b27}. }
    \label{fig:ResNet50}
\end{figure}

These later, more complex models have many more layers than the relatively simple AlexNet architecture of 5 Convolutional and 3 Fully Connected (or Dense) layers. As an example the architecture of ResNet50 shown in Fig. \ref{fig:ResNet50} provides an at-a-glance comparison with that of AlexNet. 

\subsection{CNNs in Breast Cancer Classification}

In general, research using AI for breast cancer detection and classification based on histopathology images has been conducted as a series of isolated studies arising from specific, singular datasets. These include the Bach challenge~\cite{b35}, and the BreakHis image dataset~\cite{b61}. 

\section{Related Works}
\label{sec:relatedworks}

\subsection{Table of Related Works}
Table. \ref{tab:related-works} summarises related work on the two datasets tested in this study. It is not an exhaustive list, but it represents some of the more well known studies.

\small{
\begin{table}[ht]
    \centering
    \begin{tabular} {| p{0.15\linewidth}|l| p{0.20\linewidth}  |l|p{0.15\linewidth}|}
    \hline
        \textbf{Author(s)} & \textbf{Year} & \textbf{Models} & \textbf{Data Set} & \textbf{Highest Accuracy}  \\ \hline
        Spanhol et al \cite{b72} & 2015 & AlexNet & BreakHis x40 & 90\% \\ \hline
        Bayramoglu et al \cite{b79} & 2016 & Bespoke & BreakHis x40 & 89\%  \\ \hline
        Nawaz et al \cite{b75} & 2018 & DenseNet & BreakHis x100 & 97.86\%  \\ \hline
        Deniz et al \cite{b76} & 2018 & AlexNet + VGG16 + SVM & BreakHis & 91.87\% \\ \hline
        Vidyathi et al \cite{b78} & 2019 & LeNet-5 + CLAHE + Watershed & BreakHis & 90\%  \\ \hline
        Alom et al \cite{b81} & 2019 & IRRCNN & BreakHisx100 & 97.59\% \\ \hline
        Das et al \cite{b74} & 2020 & Bespoke CNN + MIL & BreakHisx200 & 93.04\%  \\ \hline
        Mewanda et al \cite{b77} & 2020 & IRRCNN  & BreakHis & 97.58\%  \\ \hline
        Wang et al \cite{b73} & 2018 & VGG16, SVM & Bach2018 & 81.25\% \\ \hline
        Wang et al \cite{b73} & 2018 & VGG16 (only) & Bach2018 & 92.50\%  \\ \hline
        Rahklin et al \cite{b80} & 2018 & ResNet50, VGG16, InceptionV3 & Bach2018 & 87.20\%  \\ \hline
        Weiss et al \cite{b90} & 2018 & Xception, ResNet50 + VGG16 + LR & Bach2018 & 83\%  \\ \hline
        Gour et al \cite{b62} & 2020 & ResHist (ResNet based) & Bach2018 & 92.62\% \\ \hline
    \end{tabular}
    \caption{Tabulated Related Works summary} 
    \label{tab:related-works} 
\end{table}
}

\subsection{Existing comparison of works}

M. Zhu et al's 2023 paper  \cite{b60} is notable and commendable as a  resource for comparing existing CNN research in breast cancer classification. 

However, the insular nature of the underlying studies makes comparing the relative efficacy and strengths and weaknesses of the CNN models  themselves challenging. Superior performance across different studies is likely influenced more by variations in augmentation techniques, other pre-processing steps, and the innovations within the individual projects, rather than the inherent performance of the CNN models. Although the paper provides insight into the best results, it does not (and cannot) clearly identify the best models.

In addition, general research and Zhu et al.'s paper \cite{b60} highlight the absence of some contemporary CNN models that are likely well-suited for the fine-grained analysis and feature detection required in histopathology images. Notable examples include EfficientNets \cite{b28}, NASNets \cite{b32}, and MobileNetV2 \cite{b31}.

As can be seen from Table. \ref{tab:related-works}, there have been studies using ensemble models but these are in the minority. Deniz et al \cite{b76} and Weiss et al \cite{b90} both employing combinations of CNN models with traditional machine learning classifiers - SVM (Support Vector Machine) and LR (Logistic Regression) respectively. These are significant as they are conceptually similar to the ensemble models developed as part of this study. There are some studies that have considered multiple datasets, such as Mewada et al's study \cite{b77}, but in general most studies have only considered a single dataset and, if competition based, focused solely on overall classification accuracy.

\section{Aims and Contributions}
\label{sec:Aims}
This study seeks to determine the best CNN models and their optimum settings to achieve high quality results. Furthermore, to ascertain whether novel combinations of these models could provide even greater efficacy in breast cancer classification on histopathology based images. It aims to achieve this by:
\begin{itemize}
    \item \textbf{CNN Model Comparison}. Systematically comparing a range of leading standalone CNN architectures with a goal of identifying which are most suited for breast cancer classification on histopathology images. Such a detailed comparison has not yet been undertaken. This includes establishing the optimal hyperparameters, the impact of varying the complexity of fully connected layers prior to classification, and establishing the advantages and disadvantages of using existing model weights.
    \item \textbf{Augmentation Insights}. Evaluating the effect of various augmentation techniques and hyperparameter settings. 
    \item \textbf{Innovative and accurate Ensemble Architectures}. Combining high performing CNN models with complementary architectural features with diverse classifiers to construct innovative ensemble architectures, with a goal of achieving higher accuracy than their standalone counterparts.
    \item \textbf{Methodology}. Developing a framework and methodology for testing and comparing these models under standardised conditions. This includes the novel practice of pre-generating datasets to significantly expedite batched model runs and ensure identical conditions for model comparison. This, combined with auto curation of results across multiple runs for rapid analysis and comparison, facilitated the testing of a large number of permutations, in excess of 2000 model runs, to determine the best settings and thus highest quality results.
    \end{itemize}

\section{Datasets tested}
\label{sec:datasets}

This study focuses on the following two datasets. One being a more simple binary classification challenge, binary or malignant, the other being a multi classification challenge, benign, insitu, invasive or normal.

\subsection{BreakHis Dataset} 

The BreakHis dataset of malignant and benign images released by F. A. Spanhol et al in 2015 \cite{b61},  published 7909 breast cancer histopathology images of varying magnifications (x40, x100, x200, x400). The BreakHis dataset were labelled in a binary fashion: malignant or benign.
In general the class distribution between malignant and benign images is not even and is in the approximate ratio of 2.2:1, apart from the x400 magnification dataset which is in the ratio 2.1:1.

\subsection{BAch 2018 Dataset}
The Bach (BreAst Cancer Histology) dataset released as part of the Bach grand challenge on breast cancer histology images by G. Aresta et al \cite{b35} in 2018. This was a smaller dataset, 400 images, spread evenly across 4 classifications: Normal, Benign, InSitu and Invasive. The findings from the challenge can be found published in the ICIAR 2018 conference proceedings in Springer, \cite{b84}. 

The challenge also provides a separate testing dataset of 100 images. Whilst these cannot be utilised in the wider study, it is possible to use trained models and settings to predict class labels for  these images and submit them for a "blind" marking. This was undertaken in this study. The results are discussed in section \ref{sec:bach-challenge}.

\section{Models Tested}
\label{sec:Models}
\subsection{Standalone Models}

 The following standalone CNN models were evaluated:
 \begin{itemize}
     \item VGG16, VGG19  \cite{b23}
     \item InceptionV3 \cite{b54}
     \item ResNet50, ResNet152 \cite{b27}
     \item DenseNet121 \cite{b29}
     \item Xception \cite{b53}
     \item NASNetMobile  \cite{b32}
     \item MobileNetV2 \cite{b31}
     \item EfficientNetV2B1 \cite{b28}
     \item AlexNet \cite{b22} 
     \item LeNet-5 \cite{b17}
     \item Bespoke (Section \ref{sec:bespoke-architecture})
 \end{itemize}

AlexNet and LeNet were tested for historical purposes to demonstrate CNN evolution. 
In addition to industry standard models, a Bespoke Model was implemented as outlined in section \ref{sec:bespoke-architecture}.

\subsection{Bespoke Model Architecture}
\label{sec:bespoke-architecture}

In addition to using established CNN models from keras.applications, it was valuable and informative to develop a bespoke model, designed specifically for the fine grained feature extraction challenges required by medical image classification. Whilst simpler than more complex architectures such as InceptionV3 \cite{b54} or ResNet50 \cite{b27} with which it was unlikely to compete, this model, designed for efficiency and accuracy, achieved commendable performance, even outscoring more complex models such as the VGG models  \cite{b23}.

The bespoke CNN model architecture is illustrated in Fig. \ref{fig:bespoke}: 

\begin{figure}[!ht]
    \centering
    \includegraphics[width=1.0\linewidth]{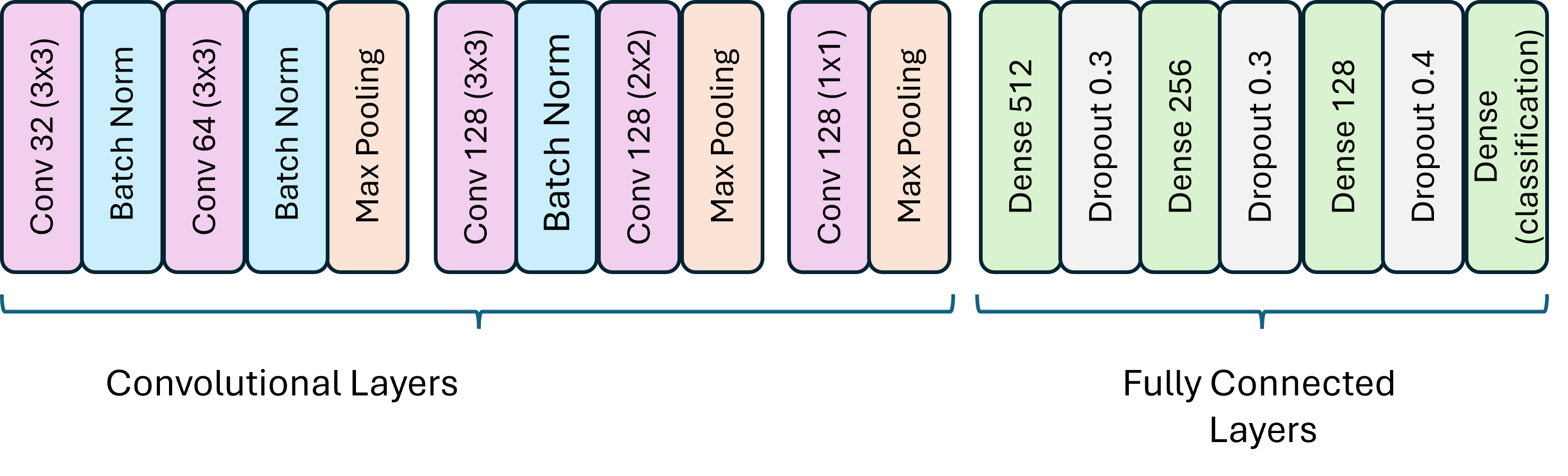}
    \caption{Bespoke CNN Model Architecture}
    \label{fig:bespoke}
\end{figure}

\subsection{Ensemble Architectures}

Based on the principle of leveraging diversity to enhance model robustness and accuracy, we subsequently integrate CNN models with distinct architectural advantages to harness their complementary strengths into ensemble architectures combined with diverse classifiers. The following CNN models are stacked together:

\begin{enumerate}
    \item DenseNet121 \cite{b29} + InceptionV3 \cite{b54} + NASNetMobile \cite{b32}

    \item EfficientNetV2B1 \cite{b28} + InceptionV3 \cite{b54} + ResNet50 \cite{b27}

    \item DenseNet121 \cite{b29} + InceptionV3 \cite{b54} + ResNet50 \cite{b27}

    \item DenseNet121 \cite{b29} + InceptionV3 \cite{b54} + MobileNetV2 \cite{b31}
   
\end{enumerate}

The stacked features from each of the three combined CNNs is married to one the following 4 classifiers, giving rise to 4 ensemble architectures per Tri-CNN model combination. The numbering in this section gives rise to the shorthand notation for the Ensembles. For example, Ensemble 1a is DenseNet121, InceptionV3, NASNetMobile + LR which is further abbreviated in some of the densely populated tables to Ens 1a.

\begin{enumerate}[label=\alph*)]
\item \textbf{Logistic Regression - LR} models the probability that an observation belongs to a specific class. One might expect strong performance on the BreakHis datasets, which involve binary classifications. However, LR's effectiveness can be sensitive to class distribution \cite{b106}, and with a ratio of approximately 2.2:1 for malignant to benign images, this imbalance could reduce its accuracy.

Despite being a multi-classification problem, the Bach dataset, with its evenly distributed classes, demonstrates the efficacy of LR when class distribution is balanced. Although scikit-learn offers a `multi\_class='multinomial'` setting, the simpler `ovr` (one-vs-rest) setting was favoured by GridSearch over multinomial for both BreakHis and Bach datasets. Thus, the model operated as a series of binary logistic regressions. GridSearch also determined that LR would employ the LBFGS solver for all datasets, a Limited-memory Broyden-Fletcher-Goldfarb-Shanno algorithm, which is effective for handling large datasets.

For a binary classification problem (`ovr` mode), where the response variable \(Y\) takes values in \(\{1, 2\}\), given predictor variables \(\mathbf{x} = (x_1, x_2, \dots, x_p)^T\), LR models the log-odds of the probabilities as follows \cite{b104}:

\begin{equation}
\log\left( \frac{P(Y=1 \mid \mathbf{x})}{P(Y=2 \mid \mathbf{x})} \right) = \beta_0 + \boldsymbol{\beta}^T \mathbf{x}
\label{eq:binary_logit}
\end{equation}

where:
\begin{itemize}
    \item \(\beta_0\) is the intercept term.
    \item \(\boldsymbol{\beta} = (\beta_1, \beta_2, \dots, \beta_p)^T\) are the coefficients corresponding to each predictor variable, where \(p\) represents the number of features (or predictor variables) in the model.
\end{itemize}

From the log-odds, the probabilities can be derived as:

\begin{equation}   
P(Y=1 \mid \mathbf{x}) = \frac{\exp(\beta_0 + \boldsymbol{\beta}^T \mathbf{x})}{1 + \exp(\beta_0 + \boldsymbol{\beta}^T \mathbf{x})}
\end{equation}

Given that \( P(Y = 2 \mid \mathbf{x}) = 1 - P(Y = 1 \mid \mathbf{x}) \), we have:

\begin{equation}
P(Y=2 \mid \mathbf{x}) =  \frac{1}{1 + \exp(\beta_0 + \boldsymbol{\beta}^T \mathbf{x})}
\end{equation}

Note: The equations provided here represent a simplified case of logistic regression, specifically for binary classification or when the multi\_class setting is set to 'ovr' (one-vs-rest). 

\item \textbf{Support Vector Classification - SVC} is designed to find the optimal hyperplane that maximizes the margin between different classes. It is capable of non-linear classification, which, achieved through the "kernel trick," enables it to discern complex patterns found in histopathology images  \cite{b102}. The kernel trick uses kernel functions to implicitly project data into a higher-dimensional space. This approach relies on calculating the dot product from the input vectors to determine the similarity between classes, allowing the algorithm to find an optimal separating hyperplane in this new feature space without directly computing the transformation.

SVC excels at higher magnifications, achieving the highest accuracy for any ensemble architecture in the BreakHis x400, and combined BreakHis datasets. It potentially handles the intricate details and textures present at higher magnifications more effectively than at lower magnifications.

Formulation of the hyperplane and margin, as defined by Cortes and Vapnik \cite{b102}, can be shown as:
\begin{equation}
\min_{\mathbf{w}, b}    \frac{1}{2} \|\mathbf{w}\|^2 + C F\left(\sum_{i=1}^{l} \xi_i\right) 
\end{equation}
subject to: $y_i (\mathbf{w}^T \phi(\mathbf{x}_i) + b) \geq 1 - \xi_i, \quad \xi_i \geq 0 $ \cite{b107}

Where:
\begin{itemize}
    \item $\mathbf{w}$ is the normal vector to the hyperplane.
    \item $b$ is the bias term in the decision function.
    \item $C$, a constant, is the regularization parameter that controls the trade-off between achieving a low training error and minimizing model complexity for better generalization.
    \item $\xi_i$ are the slack variables that allow for misclassification; larger values permit more violations of the margin. $\xi_i = 0$ for correctly classified points that lie outside or on the margin; $\xi_i > 0 $ for misclassified points (inside the margin).
    \item $\phi(\mathbf{x}_i)$ maps $\mathbf{x}_i$ to a higher-dimensional space, enabling the capture of non-linear patterns through the kernel trick. For a linear kernel, the decision function simplifies to $\mathbf{w}^T \mathbf{x}_i + b \geq 1 - \xi_i$.
    \item $F(u)$ is a monotonic convex function applied to the slack variables, designed to penalize margin violations. This function can vary; for example, a linear function $F(\xi) = \xi$ (as in the standard SVC formulation as introduced by Cortes and Vapnik \cite{b102}) simply sums the violations, whereas a quadratic function $F(\xi) = \xi^2$ penalizes larger violations more heavily. The choice of $F$ affects the SVC’s tolerance for errors and its generalization capabilities. 
\end{itemize}

A key strength of SVC is its ability to select differing kernels to optimise its classification efficacy depending on dataset. The kernel is selected as part of a grid search process using GridSearchCV. 
For each dataset run, the regularization parameter, kernel type, kernel coefficient and degree of polynomial kernel (for the poly kernel) were selected depending on the validation dataset. The Regularization parameter boundaries were pre-explored to identify the best range, in order to limit permutations and thus impact on computer resources.
Depending on which provides optimum classification efficacy, in our implementation SVC can act as a simple linear classifier or employ more complex decision boundaries using the 'rbf' or 'poly' kernels. 

\item \textbf{Random Forest - RF} is an ensemble learning method renowned for its robustness and accuracy across various applications. RF constructs multiple decision trees during training and outputs the class that is the mode of the classes predicted by individual trees. This method effectively reduces overfitting and enhances prediction accuracy, making it particularly suitable for complex classification tasks such as those found in histopathology images. \cite{b105}

RF employs the principles of bootstrap aggregation,  known as bagging, to improve model stability and accuracy. Each tree in the forest is trained on a random bootstrap sample of the data, which means each tree sees a slightly different subset of the data. This process helps in reducing variance and avoiding overfitting.
Additionally, RF utilizes the random subspace method, where each tree is not only trained on a random sample of data instances but also considers a random subset of features when making decisions. This feature randomness introduces further diversity among the trees in the forest, thereby enhancing the ensemble's ability to generalize across different datasets and reducing the likelihood of overfitting to noise present in the training data.

These characteristics make RF a powerful tool for handling datasets with complex structures and high-dimensional feature spaces, as to be found in histopathology image datasets. This is borne out in the results Table. \ref{tab:averaged-leaderboard}, which shows RF based ensembles at the top of the leaderboard. 

\item \textbf{Light Gradient Boosted Machine - LGBM} \cite{b103} is a powerful boosting algorithm that combines multiple weak classifiers (typically shallow decision trees) to form a strong predictive model. Its ability to handle imbalanced data effectively should make it particularly suitable for the BreakHis datasets, which have an imbalanced class distribution. LGBM’s weak learners incrementally correct errors from previous iterations, enabling the model to learn complex patterns in the data - as required for fine-grained feature detection in histopathology images, such as breast cancer classification.

It seemed sage to employ a boosting algorithm and LGBM is chosen over XGBoost due to its lightweight nature and efficient performance. Unlike other boosting algorithms that grow trees level-wise, LGBM grows trees leaf-wise, which  results in greater loss reduction and faster convergence. This leaf-wise approach can lead to better performance, especially on large datasets. In addition it required far less computing resources than XGBoost, which proved problematic to implement.

The update rule used by LGBM can be understood as a specific implementation of the general principles of boosting, which is a type of Forward Stagewise Additive Modelling \cite{b104}. (Note: As mentioned, LGBM distinguishes itself as a leaf-wise boosting algorithm, setting it apart from others such as XGBoost, which typically use level-wise growth.)

In each stage of the boosting process, LGBM aims to minimize the residual errors from the previous model iteration by adding a new weak learner \(h_m(x)\). The model is updated iteratively using the formula:

\begin{equation}
    F_m(x) = F_{m-1}(x) + \gamma_m h_m(x),
\end{equation}

where:
\begin{itemize}
    \item \(F_{m-1}(x)\) is the model built up to the \((m-1)\)th iteration.
    \item \(h_m(x)\) is the new weak learner (typically a decision tree) added at iteration \(m\).
    \item \(\gamma_m\) is the learning rate, controlling the contribution of \(h_m(x)\) to the overall model.
\end{itemize}

\end{enumerate}
\section{Research Methodology}
\label{sec:method-design}

\subsection{Outline Concepts}

\begin{itemize}
    \item \textbf{Data Pre-processing:} Ensuring the images are in the required format (3 channel x 8 bits/channel), images randomly split into train (60\%), validation (20\%) and test (20\%) and the directory structure reflects the image labels. 
    Post splitting the datasets, a programmatic check to ensure there is no data leakage is initiated.
    \item \textbf{Augmentation}: Pre-Training - Generating augmented images to supplement training datasets. 
    
    During Training - Comparing static augmentation (serializing datasets) with on-the-fly/dynamic augmentation. 
    \item \textbf{Dataset Pre-generation:} Serializing datasets to numpy files to ensure identical conditions for each CNN model and to expedite training. See Section \ref{sec:methodology-pregen}.
    \item \textbf{Dynamic Hyperparameters:} Enable runtime 
    adjustment of hyperparameters for flexible experimentation,
    in addition to flags that, for example, control whether to use pre-existing weights, or to adopt fine-grained filtering by employing more fully connected layers (see section \ref{sec:course-v-fine}). 
    \item \textbf{Efficient Training:} Use parameterised callbacks for early stopping, learning rate reduction on loss plateau, and saving the best model state when validation accuracy improves.

    \item \textbf{Novel ensemble architectures:} Creating ensemble models by combining high performing CNNs exhibiting complementary architectures and diverse classifiers, based on insights from the CNN model comparison.
    \item \textbf{Comprehensive Quality Metrics:} Establishing quality metrics including Accuracy, Precision, Recall, F1, and Specificity, both as overall aggregates and per class for multi-classification datasets. 
    \item \textbf{Parameter \& Quality Metrics Serialization:} Storing all parameters and hyperparameters of every CNN and ensemble run in JSON files, combined  with quality metrics from testing, to facilitate automated results analysis.
\end{itemize}

The two test cases, standalone CNN models and ensemble models, resulted in two highly parameterized code harnesses. These, together with other aspects of the methodology above, are discussed in the following sections.

\subsection{End-to-end Architecture}

\begin{figure}
    \centering
    \includegraphics[width=1\linewidth]{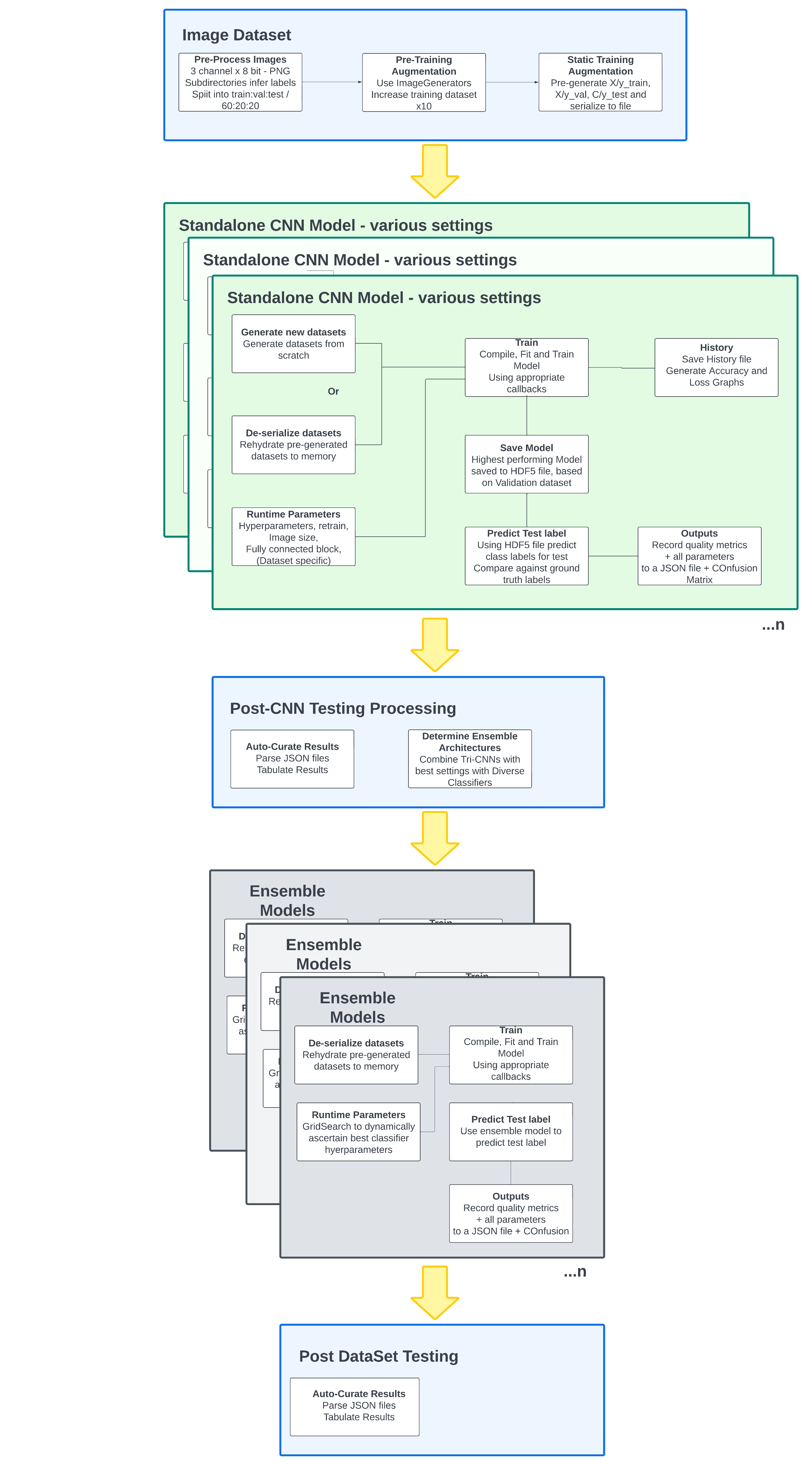}
    \caption{End-to-end Architecture - Pre-processing, CNN models in a loop, Ensemble Models in a loop - inferences}
    \label{fig:end-to-end-architecture}
\end{figure}

Fig. \ref{fig:end-to-end-architecture} outlines a simplified block architecture depicting the various stages of the methodology for each image dataset. Starting from image pre-processing, augmentation and pre-generation and serialization of datasets; through to multiple CNN model iterations based on different models and settings; through to ensemble models and post-processing.

The end-to-end method is broken down into three algorithms. 

\begin{itemize}
    \item Algorithm \ref{alg:parameter-setting}: outlines how optimum hyperparameter and other settings are established.
    \item Algorithm \ref{alg:comparison}: outlines how the main standalone CNN harness may be instantiated in batch mode.
    \item Algorithm \ref{alg:ensemble_processing}: outlines how the ensemble processing is effected.
\end{itemize}

\RestyleAlgo{ruled}

\SetKwComment{Comment}{/* }{ */}

\begin{algorithm}
\caption{Hyperparameter, Environment, and Runtime Settings}
\label{alg:parameter-setting}
\KwData{Input Image Datasets}
\KwResult{Best hyperparameters, Image Size, Fully Connected Block, Whether to retrain or use existing model weights, augmentation settings -\textgreater leading to pre-generated serialized datasets for training and optimised settings}
\For{each Image Dataset}{
    Pre-process input images - format; split into train, validation, test, with sub-directories inferring class labels\;
    Establish best settings on non-augmented data\;
    Perform electronic data leak test across directories\;
    \For{each iteration  establish the optimum settings for:  
        Initial learning rate (LR);
        LR Decay;
        Patience for Early Stopping and Reduce LR on plateau;
        Input image size;
        Retrain every layer vs. use existing weights;
        Complex fully connected block vs. simple block;
        Batch Size;}{
        Compare results, if increased accuracy and amend setting
    }
    Establish best pre-training augmentation generation settings using ImageDataGenerator\;
    \For{each iteration}{
        Rotation range, fill\_mode, width/height\_shift\_range\;
        Shear\_range, zoom\_range, horizontal/vertical\_flip\;
        Brightness\_range\;
    }
    Pre-generate training datasets using ImageDataGenerator and FlowFromDirectory, serialize to file\;
}
\end{algorithm}

\begin{algorithm}
\caption{Main Standalone Model Comparison}
\label{alg:comparison}
\KwData{Pre-Generated Image Datasets}
\KwResult{Model training run -\textgreater runtime hyperparameters and results persisted to JSON file; Model persisted to HDF5 file}
\For{each Image Dataset}{
    \For{each model in \{BespokeModel, AlexNet, LeNet-5, DenseNet121, InceptionV3, ResNet50, ResNet152, EfficientNetV2B1, NASNetMobile, MobileNetV2, Xception, VGG16, VGG19\}}{
        Deserialize datasets to memory\;
        Invoke training harness:\;
        \Indp 
        Create directories\;
        Create model\;
        Add callbacks\;
        \eIf{fine-grained filter set}{
            Add complex fully-connected pre-classifier block\;
        }{
            Add simpler fully-connected pre-classifier block\;
        }
        Compile and fit model to $X\_{\text{train}}, y\_{\text{train}}$, validating with $X\_{\text{val}}, y\_{\text{val}}$\;
        Save history file and create Accuracy and Loss by Epoch training graphs\;
        Reload best saved model from HDF5 file\;
        Predict class labels for $X\_{\text{test}}$\;
        Compare predicted labels against ground truth labels $(y\_{\text{test}})$\;
        Generate Confusion matrix and quality metrics from TP, TN, FP, FN per class\;
        Save all parameters and quality metrics per class to JSON file\;
        \Indm 
    }
}
\end{algorithm}

\begin{algorithm}
\caption{Ensemble Model Processing}
\label{alg:ensemble_processing}
\KwData{Pre-Generated Image Datasets}
\KwResult{Ensemble models, optimised parameters and results persisted to a JSON file}
\For{each Image Dataset}{
    Establish best models for Ensemble Architectures\;
    (Stack three CNN models with LR, SVC, RF, LGBM classifiers, yielding 4 ensembles)\;
    \For{each set of 3 CNN models to form an ensemble}{
        Load the models from HDF5 files up to the layer before classification (feature maps)\;
        Deserialize the datasets to memory\;
        Create feature maps for train, val, and test for each of the 3 CNN models\;
        Concatenate the feature maps from each of the models for each dataset\;
        \For{each of the 4 classifiers selected}{
            Establish best hyperparameters using GridSearch on the Validation feature map\;
            Create the classifier using the best hyperparameters\;
            Fit the classifier to the combined features of the training feature map\;
            Predict class labels against test feature map\;
            Compare predicted labels against ground truth labels $(y\_{\text{test}})$\;
            Generate Confusion matrix and quality metrics from TP, TN, FP, FN per class\;
            Save all parameters and quality metrics per class to JSON file\;
        }
    }
}
\end{algorithm}

\subsection{Data Pre-processing}
\label{sec:data-preprocessing}
For both image datasets, for each class (labelled by way of its sub-directory structure) the datasets were split -  training:validation:testing ratio of 60:20:20, using train\_test\_split from sklearn.model\_selection. Pre-processing of each dataset was slightly different as set out below: 

\subsubsection{Bach 2018 Grand Challenge}

The Bach datasets can be downloaded from zenodo \cite{b35}. The zip file ICIAR2018\_BACH\_Challenge has quite a straight forward structure with a root directory of Photos followed by sub-directories for Benign, InSitu, Invasive and Normal. Each sub-directory containing TIFF files of size 2048 x 1536 pixels, with a bit depth of 48, each approximately 18Mb in size.

Pre-processing was straightforward. The images were converted to 24bit (8 bit x3 channel) PNG files, of the same resolution but less than a quarter of the size at  approximately 4Mb each.

\subsubsection{BreakHis Datasets} 

The BreakHis datasets can be downloaded following instructions in Spanhol et al's paper \cite{b61}. The folder root directory has two sub-directories: benign and malignant.

Beneath both the benign and malignant sub-directories is a further sub-directory called SOB. Under SOB are more detailed classifications on the type of tumour, followed by the magnification level, x40, x100, x200, x400 with each of these directories containing the image files as 700 x 460 pixel, 24 bit.
The BreakHis images did not need pre-processing, the images were collated into a simpler directory structure of x40, x100, x200, x400 each with a sub-directory for benign and one for malignant, each holding the respective images. 

\subsection{Augmentation}
\label{sec:augmentation}

Augmentation of the training image datasets was an important technique, particularly for the Bach dataset, which contained only 60 images per class once split into train, val and test datasets. As a result, augmentation significantly impacted training efficacy. \\

\subsubsection{Pre-training: Augmentation To Enhance Training Datasets} To increase the training dataset size, the ImageDataGenerator function from keras.preprocessing.image was used to generate and save additional augmented images from each source image.

Careful selection of augmentation parameters was necessary to avoid over-distortion of the images. For example, a typical 'fill\_mode' of 'nearest' could create tumor-like artifacts in the augmented images. Typical parameters found online were generally not suitable. 

These parameters were used to generate 10 augmented images per original image in the training set. \\

\subsubsection{Augmentation During Training} In-place data augmentation was employed during training using a slightly different set of parameters, but again chosen with care not to distort the image features.  

For efficiency, all datasets generated during training were  serialized to numpy files for reuse. This approach standardised X\_train (as well as X\_val and X\_test) across multiple models and hyperparameter settings, ensuring that each model was trained on exactly the same dataset. 

Although serializing and reusing X\_train potentially loses some of the benefit of on-the-fly augmentation, which dynamically introduces variability to reduce overfitting, it still provided an improvement over training with the original images alone. We termed this static-augmentation at training. 

The outcomes of the augmentation approaches are detailed in section \ref{sec:results-augmentation}.

\subsection{Dataset Pre-generation}
\label{sec:methodology-pregen}

As discussed in the previous section the three datasets datasets, training, validation and testing were pre-generated and serialized to file.
This approach ensured consistent data conditions and significantly expedited model training. It allowed for consistent comparisons across more than 2000 model runs.

\subsection{CNN Model Harness}
\label{sec:cnn-model-harness}

Central to the methodology and framework is the development of a runtime configurable harness that can accept the core CNN model, the input dataset, and various runtime and model hyperparameters.

The primary benefit of this function is that it allows for multiple calls to batch test multiple models in a loop with varying runtime parameters. 
The logic of the CNN model testing harness is outlined in Algorithm \ref{alg:comparison}. 

\subsection{Coarse or Fine Grained - Fully-Connected Pre-Classification blocks}
\label{sec:course-v-fine}

Each CNN model is loaded without the very topmost layer (the classification layer) and a uniform classification construct is added. The harness tests using a simple fully connected layer block or a more complex one, prior to classification, as determined by the fine\_grained hyperparameter. 

\begin{figure}[!ht]
    \centering
    \includegraphics[width=0.8\linewidth]{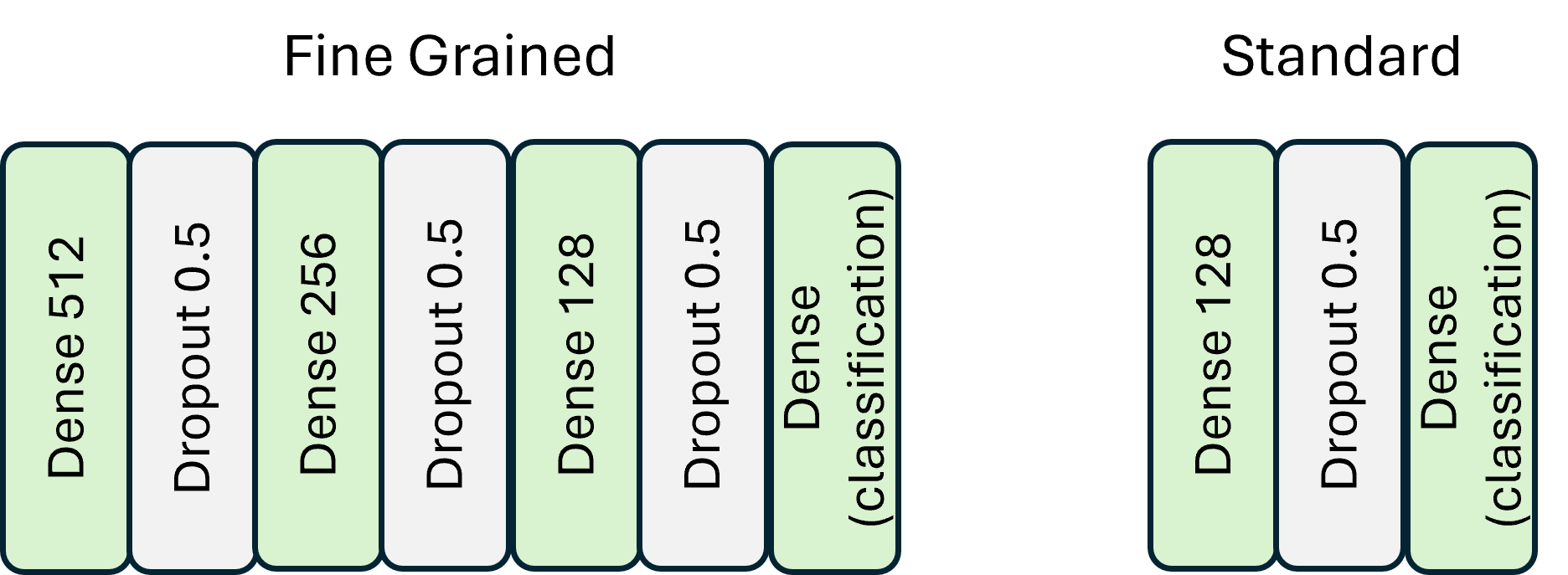}
    \caption{Dense block classification variants}
    \label{fig:DenseOptions}
\end{figure}

The two variants are shown in Fig. \ref{fig:DenseOptions}. The fine-grained model was particularly effective for the Bach dataset \cite{b35}, albeit less marked for the BreakHis binary classification datasets \cite{b61}. The impacts on the classification accuracy for the Bach dataset are detailed in Section \ref{sec:fine-v-coarse} in Table. \ref{tab:coarse-v-fine-accuracy-comparison}. 

It should be noted than the additional dense layers add significantly to the number of trainable parameters in a CNN, and thus computational resources required. As such its use was not recommended or used for the BreakHis datasets. This underlines the importance of flexible settings that may vary per image dataset characteristics.

\subsection{Ensemble Model harness}
\label{sec:methodology-ensemble-harness}

Complementing the CNN harness is another harness for the ensemble architectures. This enables the injection of either 2 or 3 model files from the CNN training  (HDF5 files) combined with the classifiers: Logistic Regression, Support Vector Classifier, Random Forest, Light Gradient Boosted Machine. 
The logic for the ensemble harness is outlined in Algorithm \ref{alg:ensemble_processing}. 

\subsection{Parameter and Quality Metrics Serialization - automated results analysis}
\label{param-and-outputs-serialization}

The hyperparameters.json file created for a standalone CNN model run and Ensemble architecture run are similar. For the CNN model runs, the JSON file captures information concerning the dataset name, whether it was augmented, whether the fine grain filter was applied, the model name and where the model file is saved; together with all the hyperparameter settings that were employed together with the results, i.e. the quality metrics, when the trained model is used to predict class labels against the test dataset. For the ensemble model runs, the specific CNN hyperparameter settings are Not Applicable, but instead the hyperparameters selected by the GridSearch process are saved to the JSON file. In every other detail the two types of JSON file are identical. This enabled automatic parsing of a range of model runs, CNN and ensemble, to facilitate rapid results analysis and cross-model comparisons.

\section{Run Outputs}

The challenge with over 2000 runs, is that it is not possible to present all the result outputs in a journal paper. The Results section, is thus very condensed and presented as a series of tables, giving various insights into models and settings efficacy in highly summarised form. In particular with so many dimensions to show, some of the key quality metrics are not shown in these summary tables. 

More detailed tables show all the outputs for a single dataset, using optimal hyperparameter, augmentation and other runtime settings - both in terms of results, confusion matrices and quality metrics. This information is replicated many hundreds of times. Specific run details can be furnished upon request.

\subsection{Standalone CNN Batch test outputs}
Each CNN model run training generates the following files:

\begin{itemize}
    \item A graph showing Validation and Training Accuracy by Epoch.
    \item A graph showing Validation and Training Loss by Epoch.
    \item A ".h5" (HDF5) file holding the best model and its weights.
    \item A JSON file storing the hyperparameters used and the results from the run. The latter are created by loading the HDF5 file and using it to predict the results from the test dataset.
    \item A confusion matrix from the test dataset predictions.
\end{itemize}

\subsubsection{Single Run Outputs}

\begin{figure}[!ht]
    \centering
    \includegraphics[width=1.0\linewidth]{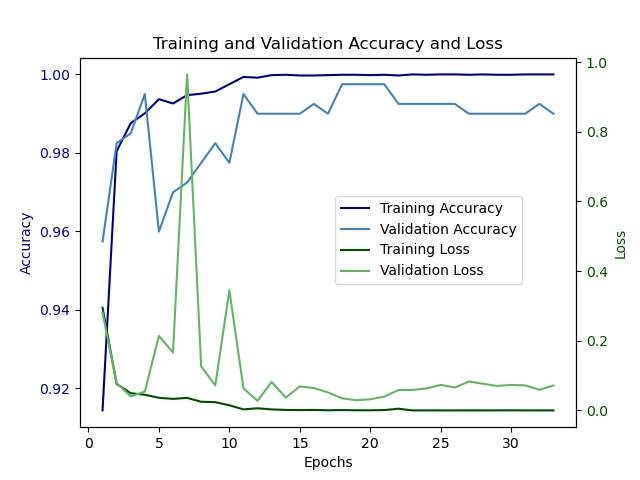}
    \caption{Training \& Validation Accuracy for DenseNet121 on BreakHisx40 }
    \label{fig:DenseNet-BreakHis40-Acc-Loss}
\end{figure}

\begin{figure}[!ht]
    \centering
    \includegraphics[width=1.0\linewidth]{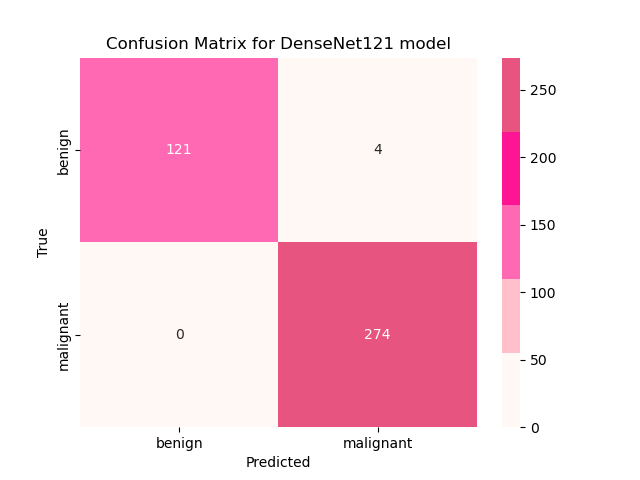}
    \caption{Confusion Matrix for DenseNet121 on BreakHis x40}
    \label{fig:DenseNet-BreakHis40-CM}
\end{figure}

The following are outputs from the BreakHis x40 magnification with augmentation and for DenseNet121. 
Fig. \ref{fig:DenseNet-BreakHis40-Acc-Loss} shows the Training and Validation Loss and Accuracy plots. Fig. \ref{fig:DenseNet-BreakHis40-CM} shows the resulting Confusion Matrix plotted when using the saved HDF5 model file to predict the classifications against the test dataset.

\subsubsection{Ensemble Run Outputs}
For an Ensemble Run, just the hyperparameters and Confusion Matrix are generated as  the training graphs shown for standalone CNN models are not applicable. The model may also be saved as a pickle file.

\subsubsection{Tabulated Results}

Having run all the CNN models and associated ensembles, we can then electronically auto-curate and tabulate the results, by parsing the directory structure and extracting the metrics from the hyperparameter.json files. Again, using the dataset BreakHis x40 in a fully retrained mode, using augmentation, an image size that preserved aspect ratio, and a range of hyperparameters, we get the results as shown in Table. \ref{tab:BreakHisx40-results} for both standalone and ensemble models. \\
\begin{table}[!ht]
\centering
\footnotesize 
\setlength{\tabcolsep}{3pt}
\renewcommand{\arraystretch}{1.0}
\begin{tabular}{|l|rrrrr|} 
\hline
\textbf{Model}   & \textbf{Acc} & \textbf{Recall} & \textbf{Prec'n} & \textbf{F1} & \textbf{Spec'ty} \\
\hline
Ensemble 1d  & 99.75\% & 100.00\% & 99.64\% & 99.82\% & 99.20\% \\
InceptionV3  & 99.50\% & 100.00\% & 99.28\% & 99.64\% & 98.40\% \\
Ensemble 1b  & 99.50\% & 100.00\% & 99.28\% & 99.64\% & 98.40\% \\
Ensemble 1c  & 99.25\% & 100.00\% & 98.92\% & 99.46\% & 97.60\% \\
Ensemble 2a  & 99.25\% & 100.00\% & 98.92\% & 99.46\% & 97.60\% \\
EfficientNetV2B1 & 99.25\%           & 99.64\%         & 99.27\%            & 99.45\%     & 98.40\%          \\
Ensemble 1a  & 99.25\% & 99.64\%  & 99.27\% & 99.45\% & 98.40\% \\
Ensemble 2b  & 99.25\% & 99.64\%  & 99.27\% & 99.45\% & 98.40\% \\
DenseNet121  & 99.00\% & 100.00\% & 98.56\% & 99.28\% & 96.80\% \\
MobileNetV2  & 99.00\% & 99.64\%  & 98.91\% & 99.27\% & 97.60\% \\
NASNetMobile & 99.00\% & 99.64\%  & 98.91\% & 99.27\% & 97.60\% \\
Ensemble 2d  & 99.00\% & 99.64\%  & 98.91\% & 99.27\% & 97.60\% \\
Ensemble 2c  & 99.00\% & 99.64\%  & 98.91\% & 99.27\% & 97.60\% \\
ResNet152    & 98.75\% & 100.00\% & 98.21\% & 99.10\% & 96.00\% \\
Xception     & 98.75\% & 99.27\%  & 98.91\% & 99.09\% & 97.60\% \\
ResNet50     & 98.50\% & 100.00\% & 97.86\% & 98.92\% & 95.20\% \\
Bespoke      & 92.23\% & 95.99\%  & 92.93\% & 94.43\% & 84.00\% \\
VGG16        & 90.73\% & 97.81\%  & 89.63\% & 93.54\% & 75.20\% \\
VGG19        & 90.23\% & 94.53\%  & 91.52\% & 93.00\% & 80.80\% \\
AlexNet      & 89.47\% & 97.08\%  & 88.67\% & 92.68\% & 72.80\% \\
LeNet        & 73.93\% & 97.81\%  & 73.22\% & 83.75\% & 21.60\% \\
\hline
\end{tabular}
\caption{Results for BreakHis x40 magnification - augmented and retrained} 
\label{tab:BreakHisx40-results} 
\end{table}

Tables such as these were created for a wide range of settings across all models; using different augmentation strategies, different pre-classification fully connected blocks, different input image sizes and ratios, and a range of hyperparameter permutations such that the very best settings could be established by comparing these tables across settings. This was a key tenet of the method and framework.

\section{Numerical Results}
\label{sec:results}

Having ascertained the very best models and settings, using optimized hyperparameters, ignoring existing model weights, using image sizes that respected the aspect ratio of the source images, augmentation and appropriate fully connected layers, it was clear that the differences between the top performing models was marginal and within the potential variation of a given training run. Despite using pre-generated datasets to ensure that images were processed in identical order and identical training augmentation, CNN model training can show slight variances between runs.

In order to more fully verify our results, we undertook more rigorous verification. This involved taking the best CNNs, with their associated optimised settings, for histopathology feature detection and cancer classification: DenseNet121 (DNet) \cite{b29}, InceptionV3 (Incp'n) \cite{b54}, ResNet50 (RNet50) and ResNet152 (RNet152) \cite{b27}, Xception (Xcep'n) \cite{b53}, NASNetMobile (NNMob) \cite{b32}, MobileNetV2 (MNet)\cite{b31}, EfficientNetV2B1 (ENet) \cite{b28}, and testing each model on each dataset ten times each to form an average. 

Ensembles 3(a,b,c,d) and 4(a,b,c,d) were also added at this time 
and each of the resulting 16 ensembles was tested a further 5 times, again to form an average. 
This verification phase of 8 CNN models and 16 ensemble models on 6 datasets, tested ten and five times respectively added an additional 960 runs.
The averaged results, which show more scientific rigour are shown in Table. \ref{tab:averaged-leaderboard}. The weighted average (WtdAvg), by which it is ordered, makes Bach equal in significance to the average of the five BreakHis (BH) datasets. 

\begin{table}[!ht]
    \centering
    \setlength{\tabcolsep}{2.5pt}
    \renewcommand{\arraystretch}{1.0}
    \begin{tabular}{|l|r|r|r|r|r|r|r|r|}
    \hline
        \textbf{Model} & BHx40 & BHx100 & BHx200 & BHx400 & BHxAll & Bach & WtdAvg\\ \hline
        Ens 4c & 99.75\% & 98.90\% & 99.75\% & 97.92\% & 98.91\% & 95.18\% & 97.11\% \\ \hline
        Ens 3c & 99.75\% & 98.90\% & 99.75\% & 97.81\% & 98.84\% & 95.18\% & 97.09\% \\ \hline
        Ens 3a & 99.50\% & 99.52\% & 99.50\% & 97.81\% & 98.23\% & 95.18\% & 97.05\% \\ \hline
        Ens 2c & 99.75\% & 98.47\% & 99.01\% & 98.03\% & 98.19\% & 95.18\% & 96.93\% \\ \hline
        Ens 3b & 99.50\% & 99.28\% & 99.50\% & 97.81\% & 98.86\% & 94.58\% & 96.78\% \\ \hline
        Ens 1c & 99.75\% & 99.14\% & 98.96\% & 98.08\% & 98.96\% & 94.58\% & 96.78\% \\ \hline
        Ens 1a & 99.50\% & 99.28\% & 98.76\% & 98.08\% & 98.74\% & 94.58\% & 96.72\% \\ \hline
        Ens 2a & 99.50\% & 98.56\% & 99.01\% & 97.26\% & 98.42\% & 94.58\% & 96.56\% \\ \hline
        Ens 4a & 99.25\% & 98.56\% & 99.50\% & 97.53\% & 98.55\% & 93.37\% & 96.03\% \\ \hline
        DNet & 98.79\% & 98.99\% & 99.11\% & 97.62\% & 98.29\% & 93.37\% & 95.97\% \\ \hline
        Ens 1b & 99.75\% & 99.28\% & 99.26\% & 98.63\% & 98.86\% & 92.77\% & 95.96\% \\ \hline
        ENet & 99.35\% & 98.37\% & 98.26\% & 96.33\% & 97.68\% & 93.86\% & 95.93\% \\ \hline
        Ens 2b & 99.75\% & 98.80\% & 99.26\% & 98.36\% & 98.61\% & 92.77\% & 95.86\% \\ \hline
        Ens 4b & 99.75\% & 99.04\% & 99.75\% & 97.81\% & 98.93\% & 92.17\% & 95.61\% \\ \hline
        Ens 2d & 99.50\% & 96.64\% & 99.01\% & 97.81\% & 97.72\% & 92.77\% & 95.45\% \\ \hline
        Xcep'n & 99.10\% & 98.61\% & 98.78\% & 96.85\% & 98.15\% & 92.53\% & 95.41\% \\ \hline
        Incp'n & 98.78\% & 98.90\% & 98.83\% & 96.96\% & 97.49\% & 92.29\% & 95.24\% \\ \hline
        Ens 4d & 99.50\% & 98.08\% & 99.50\% & 97.81\% & 98.61\% & 91.37\% & 95.03\% \\ \hline
        RNet152 & 97.99\% & 97.65\% & 97.30\% & 96.79\% & 96.17\% & 92.29\% & 94.73\% \\ \hline
        Ens 3d & 99.50\% & 98.08\% & 99.01\% & 97.81\% & 98.61\% & 90.76\% & 94.68\% \\ \hline
        RNet50 & 98.30\% & 98.32\% & 98.44\% & 97.23\% & 96.36\% & 91.33\% & 94.53\% \\ \hline
        NNMob & 98.75\% & 98.66\% & 98.04\% & 96.96\% & 98.37\% & 90.84\% & 94.50\% \\ \hline
        Ens 1d & 99.50\% & 98.08\% & 96.77\% & 97.81\% & 98.55\% & 90.76\% & 94.45\% \\ \hline
        Mnet & 99.22\% & 97.65\% & 97.42\% & 96.30\% & 97.91\% & 91.08\% & 94.39\% \\ \hline
    \end{tabular}
    \caption{Averaged results leaderboard after extensive testing}
    \label{tab:averaged-leaderboard}
\end{table}

The marginal variations between runs, underlines the importance generally of multiple CNN runs to establish which are the best performing models and settings. 
The high accuracy obtained across all models is testament to the methodology of this study, and the performance of modern architected CNNs. Ensemble Architectures were found to slightly outperform their constituent CNN models. 
\\

The following sections detail the impacts of varying the training condition settings. Where direct comparisons are made, these are for single runs, and not based on the averages set out in Table. \ref{tab:averaged-leaderboard}.

\subsection{Impacts of Augmentation}
\label{sec:results-augmentation}

This study conducted investigations on the use of augmentation as a means to boost training sample images, termed "Pre-Training Augmentation", and during training to avoid overfitting, termed "Augmentation During Training". The findings are discussed here: \\

\subsubsection{Pre-Training Augmentation}
Augmentation (section \ref{sec:augmentation}) was employed to generate 10 augmented images per original image in the training datasets. 
This significantly improved accuracy for all datasets, but the difference was particularly marked for the Bach dataset, including the blind challenge (see section \ref{sec:bach-challenge}). The results, showing the highest accuracies achieved per dataset with and without pre-augmenting the training datasets are shown in Table. \ref{tab:augmentationimpact}.

\begin{table}[!ht]
    \centering
    \begin{tabular}{|l|rr|}
    \hline
        \textbf{DataSet} & \textbf{No Augmentation} & \textbf{Augmentation} \\ \hline
        BreakHis x40 & 98.25\% & 99.75\% \\ \hline
        BreakHis x100 & 94.74\% & 99.04\% \\ \hline
        BreakHis x200 & 94.24\% & 99.50\% \\ \hline
        BreakHis x400 & 94.74\% & 97.81\% \\ \hline
        BreakHis combined & 94.74\% & 98.29\% \\ \hline
        Bach & 86.78\% & 94.58\% \\ \hline
        Bach blind test & 80.00\% & 89.00\% \\ \hline
    \end{tabular}
\caption{Comparison of Accuracy for CNN models using augmentation v no augmentation} 
\label{tab:augmentationimpact} 
\end{table} 

\subsubsection{Augmentation During Training}

The Training augmentation settings are purposely slightly different to the pre-generation augmentation settings. Training augmentation was employed in two different modes; Static and on-the-fly. The approaches and merits and drawbacks of each are discussed in section \ref{sec:augmentation}. The differences in performance between static and on-the-fly approaches is shown in Table. \ref{tab:augmentation-static-dynamic}.

\begin{table}[!ht]
    \centering
    \begin{tabular}{|l|rr|}
    \hline
        \textbf{DataSet} & \textbf{On-the-fly Augmentation} & \textbf{Static Augmentation} \\ \hline
        BreakHis x40 & 99.50\% & 99.75\% \\ \hline
        BreakHis x100 & 99.04\% & 99.04\% \\ \hline
        BreakHis x200 & 100.00\% & 99.50\% \\ \hline
        BreakHis x400 & 98.63\% & 97.81\% \\ \hline
        BreakHis combined & Not Tested & 98.29\% \\ \hline
        Bach & Not Tested & 94.58\% \\ \hline
        Bach blind test & 87.00\% & 89.00\% \\ \hline
    \end{tabular}
\caption{CNN Model Accuracy comparison for on-the-fly augmentation v static augmentation} 
\label{tab:augmentation-static-dynamic} 
\end{table} 

On-the-fly based augmentation is acknowledged to overall be superior, based on the theory of avoiding overfitting, but  the benefits seen were minimal. In addition the Bach Online challenge \cite{b35}, best results were achieved with static augmentation, see section \ref{sec:bach-challenge}. The difference in training time however is vast so when considering multiple permutations static augmentation during training represents a good compromise and its approach is advocated in this study.

\subsection{Pre-loaded model weights v Re-training}
\label{sec:preloaded-reatrined-comparison}

The impact of fully retraining a model and ignoring the existing model weights is detailed in Table. \ref{tab:using-existing-weights}. This underlines the relative uniqueness of histopathology images.

\begin{table}[ht]
    \centering
    \begin{tabular}{|l|l|l|}
    \hline
        \textbf{DataSet} & \textbf{Pre-loaded Weights} & \textbf{Re-trained} \\ \hline
        BreakHis x40 & 95.74\% & 99.75\% \\ \hline
        BreakHis x100 & 94.74\% & 99.04\% \\ \hline
        BreakHis x200 & 94.24\% & 99.50\% \\ \hline
        BreakHis x400 & 94.74\% & 97.81\% \\ \hline
        BreakHis combined & 94.74\% & 98.29\% \\ \hline
        Bach & 92.17\% & 94.58\% \\ \hline
    \end{tabular}
    \caption{A comparison of the best accuracies obtained by dataset using pre-loaded weights and retraining from scratch}
    \label{tab:using-existing-weights}
\end{table}



\subsection{Impact of Fine Grained Fully Connected Block}
\label{sec:fine-v-coarse}

The more complex fully-connected pre-classification block, termed fine grained, 
discussed in section \ref{sec:course-v-fine}, had a marked impact for the multi-class Bach dataset, but not for the binary BreakHis dataset. It was found that the architecture of this block needed to be dataset specific, and should be a design consideration when working with other datasets.

Table. \ref{tab:coarse-v-fine-accuracy-comparison} shows the best accuracies on the Bach dataset achieved with the Fine Grained and Coarse Grained pre-classification fully connected blocks respectively.

\begin{table}[!ht]
    \centering
    \setlength{\tabcolsep}{6pt}
    \renewcommand{\arraystretch}{1.0}
    \begin{tabular}{|l|c|c|} 
    \hline
    \textbf{Model} & \textbf{Fine Grained} & \textbf{Course Grained} \\ \hline
    InceptionV3 & 93.98\% & 83.73\% \\ \hline
    DenseNet121 & 92.17\% & 88.55\% \\ \hline
    MobileNetV2 & 91.57\% & 84.34\% \\ \hline
    NASNetMobile & 90.96\% & 83.13\% \\ \hline
    ResNet50 & 90.36\% & 62.65\% \\ \hline
    Xception & 90.36\% & 88.55\% \\ \hline
    ResNet152 & 86.75\% & 62.05\% \\ \hline
    EfficientNetV2B1 & 86.14\% & 62.05\% \\ \hline
    VGG16 & 83.13\% & 82.53\% \\ \hline
    VGG19 & 82.53\% & 81.93\% \\ \hline
    Bespoke & 81.93\% & 62.05\% \\ \hline
    AlexNet & 79.52\% & 62.65\% \\ \hline
    LeNet & 62.05\% & 62.65\% \\ \hline
    \end{tabular}
    \caption{Comparison of Accuracies using fine grained and coarse grained classficaton layers on standalone CNN Models on the Bach dataset \cite{b35}.}
    \label{tab:coarse-v-fine-accuracy-comparison}
\end{table}

\subsection{Bach Online blind challenge \cite{b35}}
\label{sec:bach-challenge}

Whilst without labelling it isn't possible to use the Test images in the Bach Online Challenge to compare predicted labels against ground truth labels, it is possible to submit a CSV file with labels for the test file of 100 images for a blind marking to the Bach Grand Challenge \cite{b35}. This auto-marks the file and presents the results by way of a leaderboard: https://iciar2018-challenge.grand-challenge.org/evaluation/part-a/leaderboard/. 

This provided an opportunity to benchmark the best CNN and Ensemble models on the competition site against all other entries in the competition to date. As at 27/07/2024 the leaderboard showed this paper's submission as 26th with 89\% using Ensemble 3c. Subsequently Ensemble 1b also achieved this score. There are multiple duplicates by the same authors on the leaderboard, as such this study's submission was the 5th highest by author. 

\section{Conclusions}
\label{sec:conclusions}

\label{sec:conclusions}
Systematic testing of multiple models with varying parameters and settings provided valuable insights into the relative efficacy of CNN and ensemble architectures in classifying breast cancer histopathology images, using the Bach \cite{b84} and BreakHis \cite{b61} benchmark datasets.

A novel aspect of the methodology concerned pre-generation of all data sets and serializing these to file. For the testing dataset, which had inline augmentation settings, this was termed static augmentation. This proved to be highly effective in expediting model training and ensuring consistent comparisons, forming a cornerstone of this study’s methodology. The reduction in training duration enabled multiple environment and hyperparameter permutations to be tested in a timely manner.

A simpler innovation concerned the persistence of all the run and hyperparameter settings together with the results to JSON files, such that results could be automatically processed for rapid analysis. 

Together these facilitated the identification of optimal hyperparameter settings, evaluation on the impacts of augmentation strategies, the use of differing fully connected layers and comparison of model accuracy when using existing model weights or retraining from scratch. As a result very high accuracies were achieved for the standalone CNN models. This subsequently led to creation of ensemble architecture models which attempted to refine classification accuracy further.

\appendices


\begin{thebibliography}{00}

\bibliography{bibtexrefs}











\bibitem{b10}
S. {\L}ukasiewicz, M. Czeczelewski, A. Forma, J. Baj, R. Sitarz, and
  A. Stanis{\l}awek, ``Breast cancer—epidemiology, risk factors,
  classification, prognostic markers, and current treatment strategies—an
  updated review,'' \emph{Cancers}, vol. 13, no. 17, p. 4287, 2021.


\bibitem{b12}
K. Fukushima, ``Neocognitron: A self-organizing neural network model for a
  mechanism of pattern recognition unaffected by shift in position,''
  \emph{Biological cybernetics}, vol. 36, no. 4, pp. 193--202, 1980.

\bibitem{b13}
D. H. Hubel and T. N. Wiesel, ``Receptive fields, binocular interaction and
  functional architecture in the cat's visual cortex,'' \emph{The Journal of
  physiology}, vol. 160, no. 1, p. 106, 1962.


\bibitem{b15}
Y. LeCun, B. Boser, J. S. Denker, D. Henderson, R. E. Howard, W. Hubbard, and
  L. D. Jackel, ``Backpropagation applied to handwritten zip code
  recognition,'' \emph{Neural Computation}, vol. 1, no. 4, pp. 541--551, 1989.


\bibitem{b17}
Y. Lecun, L. Bottou, Y. Bengio, and P. Haffner, ``Gradient-based learning
  applied to document recognition,'' \emph{Proceedings of the IEEE}, vol. 86,
  no. 11, pp. 2278--2324, 1998.


\bibitem{b19}
O. Russakovsky, J. Deng, H. Su, J. Krause, S. Satheesh, S. Ma, Z. Huang,
  A. Karpathy, A. Khosla, M. Bernstein, A. C. Berg, and L. Fei-Fei, ``{ImageNet
  Large Scale Visual Recognition Challenge},'' \emph{International Journal of
  Computer Vision (IJCV)}, vol. 115, no. 3, pp. 211--252, 2015.

\bibitem{b22}
A. Krizhevsky, I. Sutskever, and G. E. Hinton, ``Imagenet classification with
  deep convolutional neural networks,'' \emph{Advances in neural information
  processing systems}, vol. 25, 2012.

\bibitem{b23}
K. Simonyan and A. Zisserman, ``Very deep convolutional networks for
  large-scale image recognition,'' \emph{arXiv preprint arXiv:1409.1556}, 2014.


\bibitem{b25}
C. Szegedy, W. Liu, Y. Jia, P. Sermanet, S. Reed, D. Anguelov, D. Erhan,
  V. Vanhoucke, and A. Rabinovich, ``Going deeper with convolutions,'' in
  \emph{Proceedings of the IEEE conference on computer vision and pattern
  recognition}, 2014, pp. 1--9.


\bibitem{b27}
K. He, X. Zhang, S. Ren, and J. Sun, ``Deep residual learning for image
  recognition,'' in \emph{Proceedings of the IEEE conference on computer vision
  and pattern recognition}, 2016, pp. 770--778.

\bibitem{b28}
M. Tan and Q. Le, ``Efficientnet: Rethinking model scaling for convolutional
  neural networks,'' in \emph{International conference on machine
  learning}.\hskip 1em plus 0.5em minus 0.4em\relax PMLR, 2019, pp. 6105--6114.

\bibitem{b29}
G. Huang, Z. Liu, L. Van Der~Maaten, and K. Q. Weinberger, ``Densely connected
  convolutional networks,'' in \emph{Proceedings of the IEEE conference on
  computer vision and pattern recognition}, 2017, pp. 4700--4708. 


\bibitem{b31}
M. Sandler, A. Howard, M. Zhu, A. Zhmoginov, and L.-C. Chen, ``Mobilenetv2:
  Inverted residuals and linear bottlenecks,'' in \emph{Proceedings of the IEEE
  conference on computer vision and pattern recognition}, 2018, pp. 4510--4520.

\bibitem{b32}
B. Zoph, V. Vasudevan, J. Shlens, and Q. V. Le, ``Learning transferable
  architectures for scalable image recognition,'' in \emph{Proceedings of the
  IEEE conference on computer vision and pattern recognition}, 2018, pp.
  8697--8710.


\bibitem{b34}
C. Carr, F. Kitamura, G. Partridge, inversion, J. Kalpathy-Cramer, J. Mongan,
  K. Andriole, Lavender, M. Vazirabad, M. Riopel, R. Ball, S. Dane, and
  Y. Chen, ``Rsna screening mammography breast cancer detection,'' 2022.

\bibitem{b35}
G. Aresta, T. Ara{\'u}jo, S. Kwok, S. S. Chennamsetty, M. Safwan, V. Alex,
  B. Marami, M. Prastawa, M. Chan, M. Donovan \emph{et~al.}, ``Bach: Grand
  challenge on breast cancer histology images,'' \emph{Medical image analysis},
  vol. 56, pp. 122--139, 2019.

\bibitem{b46}
{International Agency for Research on Cancer - W.H.O}. (2024) Press release no.
  345. [Online]. Available:

\bibitem{b53}
F. Chollet, ``Xception: Deep learning with depthwise separable convolutions,''
  in \emph{Proceedings of the IEEE conference on computer vision and pattern
  recognition}, 2017, pp. 1251--1258.

\bibitem{b54}
C. Szegedy, V. Vanhoucke, S. Ioffe, J. Shlens, and Z. Wojna, ``Rethinking the
  inception architecture for computer vision,'' in \emph{Proceedings of the
  IEEE conference on computer vision and pattern recognition}, 2016, pp.
  2818--2826.

\bibitem{b60}
Z. Zhu, S.-H. Wang, and Y.-D. Zhang, ``A survey of convolutional neural network
  in breast cancer,'' \emph{Computer modeling in engineering \& sciences:
  CMES}, vol. 136, no. 3, p. 2127, 2023.

\bibitem{b61}
F. A. Spanhol, L. S. Oliveira, C. Petitjean, and L. Heutte, ``A dataset for
  breast cancer histopathological image classification,'' \emph{Ieee
  transactions on biomedical engineering}, vol. 63, no. 7, pp. 1455--1462,
  2015.

    
\bibitem{b62}
M. Gour, S. Jain, and T. Sunil~Kumar, ``Residual learning based cnn for breast
  cancer histopathological image classification,'' \emph{International Journal
  of Imaging Systems and Technology}, vol. 30, no. 3, pp. 621--635, 2020.

\bibitem{b72}
F. A. Spanhol, L. S. Oliveira, C. Petitjean, and L. Heutte, ``Breast cancer
  histopathological image classification using convolutional neural networks,''
  in \emph{2016 International Joint Conference on Neural Networks (IJCNN)},
  2016, pp. 2560--2567.

\bibitem{b73}
Y. Wang, L. Sun, K. Ma, and J. Fang, ``Breast cancer microscope image
  classification based on cnn with image deformation,'' in \emph{Image Analysis
  and Recognition: 15th International Conference, ICIAR 2018, P{\'o}voa de
  Varzim, Portugal, June 27--29, 2018, Proceedings 15}.\hskip 1em plus 0.5em
  minus 0.4em\relax Springer, 2018, pp. 845--852.

\bibitem{b74}
K. Das, S. Conjeti, J. Chatterjee, and D. Sheet, ``Detection of breast cancer
  from whole slide histopathological images using deep multiple instance cnn,''
  \emph{IEEE Access}, vol. 8, pp. 213\,502--213\,511, 2020.

\bibitem{b75}
M. Nawaz, A. A. Sewissy, and T. H. A. Soliman, ``Multi-class breast cancer
  classification using deep learning convolutional neural network,''
  \emph{International Journal of Advanced Computer Science and Applications},
  vol. 9, no. 6, 2018. [Online]. Available:
  {http://dx.doi.org/10.14569/IJACSA.2018.090645}

\bibitem{b76}
E. Deniz, A. {\c{S}}eng{\"u}r, Z. Kadiro{\u{g}}lu, Y. Guo, V. Bajaj, and
  {\"U}. Budak, ``Transfer learning based histopathologic image classification
  for breast cancer detection,'' \emph{Health information science and systems},
  vol. 6, pp. 1--7, 2018.

\bibitem{b77}
H. K. Mewada, A. V. Patel, M. Hassaballah, M. H. Alkinani, and K. Mahant,
  ``Spectral--spatial features integrated convolution neural network for breast
  cancer classification,'' \emph{Sensors}, vol. 20, no. 17, p. 4747, 2020.

\bibitem{b78}
A. Vidyarthi, J. Shad, S. Sharma, and P. Agarwal, ``Classification of breast
  microscopic imaging using hybrid clahe-cnn deep architecture,'' in \emph{2019
  Twelfth International Conference on Contemporary Computing (IC3)}.\hskip 1em
  plus 0.5em minus 0.4em\relax IEEE, 2019, pp. 1--5.

\bibitem{b79}
N. Bayramoglu, J. Kannala, and J. Heikkilä, ``Deep learning for magnification
  independent breast cancer histopathology image classification,'' in
  \emph{2016 23rd International Conference on Pattern Recognition (ICPR)},
  2016, pp. 2440--2445.

\bibitem{b80}
A. Rakhlin, A. Shvets, V. Iglovikov, and A. A. Kalinin, ``Deep convolutional
  neural networks for breast cancer histology image analysis,'' in \emph{Image
  Analysis and Recognition: 15th International Conference, ICIAR 2018,
  P{\'o}voa de Varzim, Portugal, June 27--29, 2018, Proceedings 15}.\hskip 1em
  plus 0.5em minus 0.4em\relax Springer, 2018, pp. 737--744.

\bibitem{b81}
M. Z. Alom, C. Yakopcic, M. S. Nasrin, T. M. Taha, and V. K. Asari, ``Breast
  cancer classification from histopathological images with inception recurrent
  residual convolutional neural network,'' \emph{Journal of digital imaging},
  vol. 32, pp. 605--617, 2019.

\bibitem{b84}
A. Campilho, F. Karray, and B. ter Haar~Romeny, ``Preface: Image analysis and
  recognition: 15th international conference, iciar 2018, p\`ovoa de varzim,
  portugal, june 27-29, 2018, proceedings,'' \emph{Lecture Notes in Computer
  Science}, vol. 10882, pp. 715--940, 2018. 

\bibitem{b88}
C. Duggan, A. Dvaladze, A. F. Rositch, O. Ginsburg, C.-H. Yip, S. Horton,
  R. Camacho~Rodriguez, A. Eniu, M. Mutebi, J.-M. Bourque \emph{et~al.}, ``The
  breast health global initiative 2018 global summit on improving breast
  healthcare through resource-stratified phased implementation: methods and
  overview,'' \emph{Cancer}, vol. 126, pp. 2339--2352, 2020.


\bibitem{b90}
N. Weiss, H. Kost, and A. Homeyer, ``Towards interactive breast tumor
  classification using transfer learning,'' in \emph{Image Analysis and
  Recognition: 15th International Conference, ICIAR 2018, P{\'o}voa de Varzim,
  Portugal, June 27--29, 2018, Proceedings 15}.\hskip 1em plus 0.5em minus
  0.4em\relax Springer, 2018, pp. 727--736.

\bibitem{b102}
C. Cortes and V. Vapnik, Support-vector networks. Mach Learn 20, 273–297. 1995. https://doi.org/10.1007/BF00994018

\bibitem{b103}
G. Ke, Q. Meng, T. Finley, T. Wang, W. Chen, W. Ma, Q. Ye and T. Y. Liu, 2017. Lightgbm: A highly efficient gradient boosting decision tree. Advances in neural information processing systems, 30

\bibitem{b104}
J. Friedman, T. Hastie and R. Tibshirani (2001). The Elements of Statistical Learning. Springer Series in Statistics Ch 10. DOI: 10.1007/b94608

\bibitem{b105}
L. Breiman.  Random forests. Machine learning, 45, pp.5-32. 2001

\bibitem{b106} A. H. A. Rahim, N. A. Rashid, A. Nayan, and A. Ahmad, "Smote approach to imbalanced dataset in logistic regression analysis," in *Proc. 3rd Int. Conf. Comput., Math., Statist. (iCMS2017) Transcending Boundaries, Embracing Multidisciplinary Diversities*, Springer, 2019, pp. 429-433.

\bibitem{b107} A. Ng, "CS229: Machine Learning - Support Vector Machines," Stanford University, 2020. [Online]. Available: https://cs229.stanford.edu/notes2020spring/cs229-notes3.pdf


\end{thebibliography}
\end{document}